\documentclass{article}

\usepackage[nonatbib,preprint]{neurips_2021}

\usepackage[utf8]{inputenc} % allow utf-8 input
\usepackage[T1]{fontenc}    % use 8-bit T1 fonts
\usepackage{hyperref}       % hyperlinks
\usepackage{url}            % simple URL typesetting
\usepackage{booktabs}       % professional-quality tables
\usepackage{amsfonts}       % blackboard math symbols
\usepackage{amsmath}
\usepackage{amssymb}
\usepackage{nicefrac}       % compact symbols for 1/2, etc.
\usepackage{microtype}      % microtypography
\usepackage{xcolor}         % colors
\usepackage{graphicx}
\usepackage{caption}
\usepackage{subcaption}

\title{Differentiable Random Access Memory using Lattices}

\author{%
  Adam P. Goucher\\
  Department of Pure Mathematics and Mathematical Statistics, University of Cambridge\\
  \texttt{goucher@dpmms.cam.ac.uk}\\
  \And
  Rajan Troll\\
  InBalance\\
  \texttt{rajan@inbalanceresearch.com} \\
}

\begin{document}

\maketitle

\begin{abstract}
  We introduce a differentiable random access memory module with $O(1)$ performance regardless of size, scaling to billions of entries. The design stores entries on points of a chosen lattice to calculate nearest neighbours of arbitrary points efficiently by exploiting symmetries. Augmenting a standard neural network architecture with a single memory layer based on this, we can scale the parameter count up to memory limits with negligible computational overhead, giving better accuracy at similar cost. On large language modelling tasks, these enhanced models with larger capacity significantly outperform the unmodified transformer baseline. We found continued scaling with memory size up to the limits tested.
\end{abstract}

\section{Introduction}

Deep Neural Networks (DNNs) have shown strong scaling with size in several domains, especially those with large amounts of available data such as NLP. Standard architectures have linear cost scaling with parameter count, requiring huge computational budgets as they scale. In particular, they have no way of recalling knowledge except a linear scan over all of their parameters.

With model sizes growing into the hundreds of billions \cite{gpt3}, recently proposed dynamic sparse models such as product key memories (PKM)\cite{pkm} and sparse mixture-of-experts (SMoE)\cite{smoe} \cite{switch} have shown good performance on these high-data domains. However, even these have asymptotic scaling $O(\sqrt{N})$ with the number of parameters, far better than linear but much worse than conventional data structures such as hashtables and trees.

\begin{figure}
  \centering
  \includegraphics[width=0.5\textwidth]{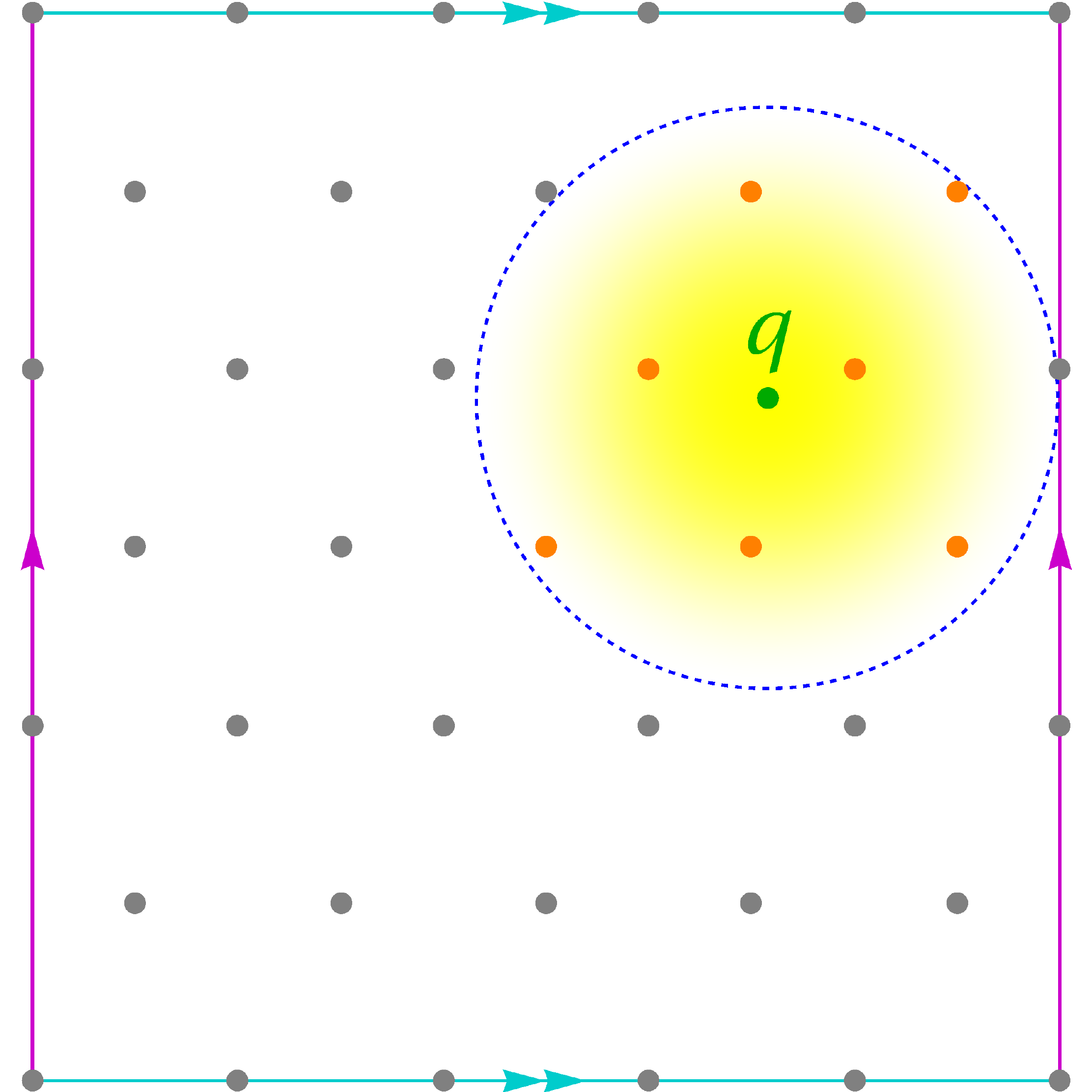}
  \caption{Example of lookup in the analogous 2D hexagonal lattice, showing the lattice points used to interpolate the query value as well as the global torus structure.}
  \label{fig:hexagonal}
\end{figure}

In this work, we introduce a lattice-based differentiable random-access memory layer (LRAM), which provides look-up functionality in $O(1)$ regardless of the number of parameters in the layer. We use a type of sparse attention to achieve this, storing values at each lattice point. Given a query point, we lookup nearby lattice points in constant time by taking advantage of the symmetry of the lattice, and then interpolate the values with a Gaussian-like kernel function to produce the output. Figure~\ref{fig:hexagonal} illustrates this process with a hexagonal lattice in two dimensions; in practice we use the 8-dimensional $E_8$ lattice.

To test the new layer, we focus on language modelling as a high-data task that scales well with model size. In particular, we expect all of our models to underfit, an ideal use-case for high-capacity sparse layers. We take a standard BERT-style\cite{bert} transformer\cite{aiayn} architecture, and replace one of the fully-connected layers with a LRAM layer of varying size, trained with a masked language modelling objective. We see strong performance improvements over the baseline for the same computational cost, and steady scaling with memory size up to the largest sizes tested. Overall:

\begin{itemize}
\item We introduce a new layer with large capacity but fast train and inference time;
\item Our fast indexing uses lattice symmetries to find the relevant parameters in constant time, regardless of layer size, assuming we have random access over the parameter storage;
\item We demonstrate performance on medium-sized transformers, outperforming the baseline with mild runtime increase.
\end{itemize}

\section{The LRAM Module}

\subsection{Overview}

We design a memory module based on a lookup function from $\mathbb{R}^n \rightarrow \mathbb{R}^m$, with $n$ determined by the chosen lattice dimension and $m$ chosen freely. To perform a lookup, we find nearby lattice points, each of which holds a parameter vector in $\mathbb{R}^m$, and then interpolate between the vectors, weighted by a kernel function favouring closer points. Each lookup only requires accessing the parameters at a small, constant number of nearby points, found in constant time, whereas the lattice can store values at arbitrarily many points and thus have a large capacity. In order to limit the number of points in the lattice, we wrap space around after a certain length in each dimension; Figure~\ref{fig:hexagonal} shows example wrapping in 2D. To adjust the model capacity, we control the number of lattice points within the wrapped space. In practice, we do multiple parallel lookups into the memory, and use learnable dense linear layers before and after, as well as a specialised activation layer to generate query locations on the torus.

\subsection{Mathematical Formulation}

We start by defining, for an $n$-tuple $K = (K_1, \dots, K_n)$ of positive
reals, the lattice:

$$ L_K := \prod\limits_{i=1}^n (K_i \mathbb{Z}) $$

and the associated torus:

$$ T_K := \mathbb{R}^n / L_K
= \prod\limits_{i=1}^n \mathbb{R} / (K_i \mathbb{Z}) $$

The torus $T_K$ inherits a quotient metric $d$ from the Euclidean metric on
$\mathbb{R}^n$.

For any superlattice $\Lambda \supseteq L_K$, we can consider $M = \Lambda /
L_K$ as a (finite) subset of $T_K$. We refer to the elements of our finite
set $M$ as \emph{memory locations}. Suppose that, at each memory location
$k \in M$, we have a fixed `value vector' $v_k \in \mathbb{R}^m$. Then, we
can define the function $\varphi : T_K \rightarrow \mathbb{R}^m$ as follows:

$$ \varphi(q) = \sum\limits_{k \in M} f(d(q, k)) v_k $$

where $f : \mathbb{R} \rightarrow \mathbb{R}$ is a monotone-decreasing
function of the distance $d(q, k)$. Although this looks like a sum over the entire memory, requiring linear time, in practice we choose our kernel $f$ such that only nearby points have nonzero effect, producing high input-dependent sparsity.

\subsection{Activation Layer for Parameterising the Torus}

In order to use this lookup function inside a neural network, we need some
method of converting an input vector into a point on the torus $T_K$.
We initially tried a direct embedding with the quotient map,
in this case modulo, but preliminary runs found poor memory utilisation and performance, so we
choose to embed the torus in $\mathbb{C}^n$ as a product of $n$ unit circles,
using the arguments of the (complex) entries of the input vector to
parameterise the torus $T_K$.

We avoid the discontinuity in $\arg{z}$ at the origin $z = 0$ by scaling
the output of the lookup function by the harmonic mean of the absolute
values of the $n$ entries. In addition to ensuring (Lipschitz) continuity
of the resulting function $\vartheta : \mathbb{C}^n \rightarrow \mathbb{R}^m$,
this has the auxiliary benefit of enabling the neural network to control
the magnitude of the output of the lookup: we have $\vartheta(\lambda z) =
\lambda \vartheta(z)$ for all non-negative real scalars $\lambda \geq 0$,
so increasing the magnitude of the input vector $z$ proportionally increases
the magnitude of the output vector.

To summarise, our neural network layer is the function:

$$ \vartheta(z_1, z_2, \dots, z_n) := \left( \dfrac{1}{|z_1|} +
\dfrac{1}{|z_2|} + \cdots + \dfrac{1}{|z_n|} \right)^{-1}
\varphi \left( \dfrac{K_1}{2 \pi} \arg{z_1}, \dfrac{K_2}{2 \pi} \arg{z_2},
\dots, \dfrac{K_n}{2 \pi} \arg{z_n} \right) $$

By identifying $\mathbb{C}^n$ with $\mathbb{R}^{2n}$, this can be regarded
as a neural network layer with $2n$-dimensional inputs and $m$-dimensional
outputs. In practice, we have $h$ heads operating in parallel on the same
memory, giving a layer from $\mathbb{R}^{2hn}$ to $\mathbb{R}^{hm}$

\subsection{Choice of Lattice}

For all of our experiments, we take $n = 8$ and let $\Lambda$ be a scaled
copy of the 8-dimensional $E_8$ lattice. This lattice enjoys a number of
favourable properties, such as being the densest way to arrange disjoint
unit spheres in 8-dimensional space \cite{viazovska} and possessing a large rotational
symmetry group. Compared to the $\mathbb{Z}^8$ lattice, this allows a far higher sparsity
of access while preserving good coverage of the space: we take the kernel radius as
$\sqrt{2}$ times the covering radius, to ensure that any query lies well within the
interpolation neighbourhood of a lattice point, and determine the maximum
number of lattice points any query requires, as well as the average across a
uniform distribution. The results are summarised in Table~\ref{tab:lattices}.

\begin{table}
\caption{Comparison between different lattices in 8 and higher dimensions. The
entries marked (m.c.) were estimated using Monte Carlo with at least 10 million
samples; everything else was determined analytically. All lattices are scaled
to be unimodular (determinant 1) so that the packing and covering radii are
directly comparable between lattices of the same dimension.}
\label{tab:lattices}
\centering
\begin{tabular}{c c c c c c}
  \toprule
  Lattice & $\mathbb{Z}^8$ & $E_8$ & $K_{12}$ & $\Lambda_{16}$ & $\Lambda_{24}$ \\
  \midrule
  Dimension & 8 & 8 & 12 & 16 & 24 \\
  Determinant & 1 & 1 & 1 & 1 & 1 \\
  Packing radius & 0.5 & 0.707 & 0.760 & 0.841 & 1 \\
  Covering radius & 1.414 & 1 & 1.241 & 1.456 & 1.414 \\
  \midrule
  Minimum points in kernel support & 768 (m.c.) & 45 (m.c.) & & & \\
  Average points in kernel support & 1039 & 64.94 & 1138 & 24704 & 32373 \\
  Maximum points in kernel support & 1312 (m.c.) & 121 & & & \\
  \bottomrule
\end{tabular}
\end{table}

$E_8$ has the same density as the cubic lattice $\mathbb{Z}^8$, but has a
higher packing radius and lower covering radius \cite{splag}. This means that
the lattice points themselves are well separated, but every point in the
ambient space $\mathbb{R}^8$ is reasonably close to at least one lattice
point. Compared to $\mathbb{Z}^8$, lookup with $E_8$ accesses 16 times fewer
points on average for the same spatial resolution.

We choose $n = 8$ rather than a higher dimension as the
higher-dimensional lattices (such as the Coxeter-Todd lattice $K_{12}$,
Barnes-Wall lattice $\Lambda_{16}$, and Leech lattice $\Lambda_{24}$) all
involve a much larger average number of points per lookup as shown in
Table~\ref{tab:lattices}.

For convenience, we set the scale of the lattice $\Lambda$ to be double that
of the conventional (unimodular) $E_8$ lattice, such that all lattice points
in $\Lambda$ have integer coordinates. In particular, $\Lambda$ has a packing
radius of $\sqrt{2}$, a covering radius of $2$, and is defined as the set of
integer coordinates with constant parity and coordinate sum divisible by 4:

$$ \Lambda := \left\{ x \in (2 \mathbb{Z})^8 \cup (2 \mathbb{Z} + 1)^8
\textrm{ such that } \sum\limits_{i=1}^8 x_i \equiv 0 \mod 4 \right\} $$

\subsection{Choice of Kernel}

We use an approximately Gaussian symmetric kernel supported only within a
certain radius, $f(r) = \max\left(0, 1 - \frac{1}{8} r^2 \right)^4$, which
means that the value $\varphi(q)$ depends only on the keys $k$ within a
distance less than $\sqrt{8}$ from the query vector $q$. Since the distance
between each lattice point and its nearest neighbours is $\sqrt{8}$, we have
$\varphi(k) = v_k$ for each lattice point $k \in M$; in other words, $\varphi$
can be considered to be smoothly interpolating between the values at the
lattice points.

This function was chosen such that the `total weight' $\sum\limits_{k \in M}
f(d(q, k))$ is equal to 1 both at the lattice points and the \emph{deep holes}
\cite{splag} maximally distant from the lattice points. For an arbitrary point $x$, the
total weight satisfies $0.851 \approxeq \frac{22158 - 625 \sqrt{5}}{24389}
\leq w(x) \leq 1$, so we did not need to normalise weightings.

\subsection{Implementation}

Given a query point $q \in \mathbb{R}^8$, we need to be able to efficiently
determine all lattice points $k \in \Lambda$ satisfying $\lVert q - k \rVert
< \sqrt{8}$.

Our solution to this problem is to apply an isometry $\phi$ of the lattice
which maps $q$ into a small \emph{fundamental region} $F \subseteq
\mathbb{R}^8$. In particular, the isometries that we consider are the
compositions of:

\begin{itemize}
\item Translation by a vector in the lattice $\Lambda$;
\item Arbitrary permutations of the eight coordinates ($8!$ choices);
\item Sign changes in an even number of coordinates ($2^7$ choices).
\end{itemize}

As such, we can guarantee that the image $\phi(q)$ is closer to the origin
than to any other lattice point, that the coordinates are in descending
order in absolute value, and that the first seven coordinates are non-negative.
That is to say, the fundamental region $F$ is defined by the constraints:

\begin{itemize}
\item $z_1 \geq z_2 \geq z_3 \geq z_4 \geq z_5 \geq z_6 \geq z_7 \geq |z_8|$;
\item $z_1 + z_2 \leq 2$;
\item $z_1 + z_2 + z_3 + z_4 + z_5 + z_6 + z_7 + z_8 \leq 4$.
\end{itemize}

By means of convex quadratic programming, we determined that there are
exactly 232 lattice points within a distance of less than $\sqrt{8}$ of
the nearest point in $F$. We store these 232 points in a fixed precomputed
array and apply the inverse isometry $\phi^{-1}$ to these points to obtain
(a superset of) all of the lattice points within the open ball of radius
$\sqrt{8}$ centred upon $q$.

Note that the isometries we consider do not generate the full isometry group
of the lattice $\Lambda$, but rather form an index-135 maximal subgroup.
Nonetheless, we use this smaller group for simplicity: we can represent $\phi$
as a composition of a translation with a signed permutation, rather than as
a general affine transformation, and therefore implement $\phi^{-1}$ more
efficiently than using a general matrix-vector multiplication.

We have implemented this procedure as a CUDA kernel which, for each query
point $q \in \mathbb{R}^8$, outputs the set of indices $\alpha_i$ of the
nearest lattice points together with their corresponding weights $w_i :=
f(d(q, k_{\alpha_i}))$ and the partial derivatives $\partial w_i / \partial
q_j$. This kernel is called by an autograd-compatible PyTorch wrapper allowing
use within a neural network during training and inference.

In practice, to speed up the layer by reducing the memory bandwidth requirements,
as in \cite{pkm}, we restrict the final weighted average to only the closest $k=32$
lattice points. Using 100 million uniformly distributed queries, we Monte Carlo
estimate that these 32 points contain on average 99.5\% of the total weight, and
at least 90\%.

\section{Methodology}

\subsection{Architecture}

To test the LRAM layer, we incorporate it into a BERT-style \cite{bert}
transformer model consisting of a residual tower of alternating fully-connected
subnetworks (applied to each element of the sequence) and self-attention modules.
Specifically, we replace one of the fully-connected subnetworks with a
memory-augmented subnetwork comprising:

\begin{itemize}
\item a dense affine layer $\mathbb{R}^w \rightarrow \mathbb{R}^w$;
\item our custom memory layer $\vartheta : \mathbb{R}^w \rightarrow \mathbb{R}^{4w}$
with $(n, m, h) = (8, 64, \frac{w}{16})$;
\item a dense affine layer $\mathbb{R}^{4w} \rightarrow \mathbb{R}^w$.
\end{itemize}

That is to say, we replace the fixed elementwise nonlinear activation
function with a trainable blockwise nonlinear function $\vartheta$, and
modify the output dimension of the input dense affine layer accordingly
(reducing it from $4w$ to $w$).

\subsection{Hyperparameters}
\label{sub:hypers}
All models were trained with Adam \cite{adam} with constant learning rate of $10^{-4}$ for normal parameters and $10^{-3}$ for memory layer parameters to compensate for sparse access. For the entire training duration, in all models, notable progress continued so we did not decrease the learning rate. The reported experiments did not use dropout we found it detrimental to the model training.

We used a 6-layer transformer with a width (bottleneck dimension) of $w = 512$, maximum sequence length of 256 tokens, and batch size of 32. The dense fully-connected subnetworks used a hidden width of 2048 with GELU activation \cite{gelu}. The memory layer, if present, replaces the fully-connected subnetwork in the fourth transformer layer.

\subsection{Dataset}
\label{sub:dataset}

Our dataset consists of paragraphs from the English Wikipedia (14 GB)
\cite{enwiki}, a book corpus (6 GB) \cite{books1} (based on \cite{books2}
and \cite{books3}), and OpenWebText (40 GB) \cite{webtext}.

The full dataset was randomly shuffled at the paragraph granularity
and separated into training (227.4M paragraphs), validation (25k
paragraphs), and testing (25k paragraphs) subsets.

The resulting dataset was preprocessed using the same pipeline as in the
XLM repository \cite{xlm}, including the conversion to lowercase and the
removal of diacritical marks. We used the byte pair encoding (BPE) algorithm
to tokenise the data with a dictionary size of 30k.

\section{Results}

\subsection{Language Modelling}
We test our sparse memory layer on masked language modelling, a domain with tons of data known to benefit from very large parameter counts. We compare against a baseline of a standard transformer, as well as against a transformer using PKM (with 8 heads, $N = 2^{16}$, value dimension 512, and key dimension 64) instead of LRAM. As it was found to be the best configuration in \cite{pkm}, we used batch normalisation for the query vectors in both the LRAM and PKM experiments.

With large enough memory capacities, our lattice-based memory layer outperforms the baseline in terms of perplexity, with only minor speed degradation. We observe little slowdown with increasing size, as predicted by our $O(1)$ scaling in a random-access model. Our augmented transformers do not match the perplexity of the PKM-augmented transformer, but have faster runtimes. At larger scales, we expect the speed advantage to grow due to the difference in complexity growth rates; but in a memory-constrained scenario, PKM-based approaches with their higher lookup dimension seem to do better.

\begin{figure}
  \centering
  \includegraphics[width=0.98\textwidth]{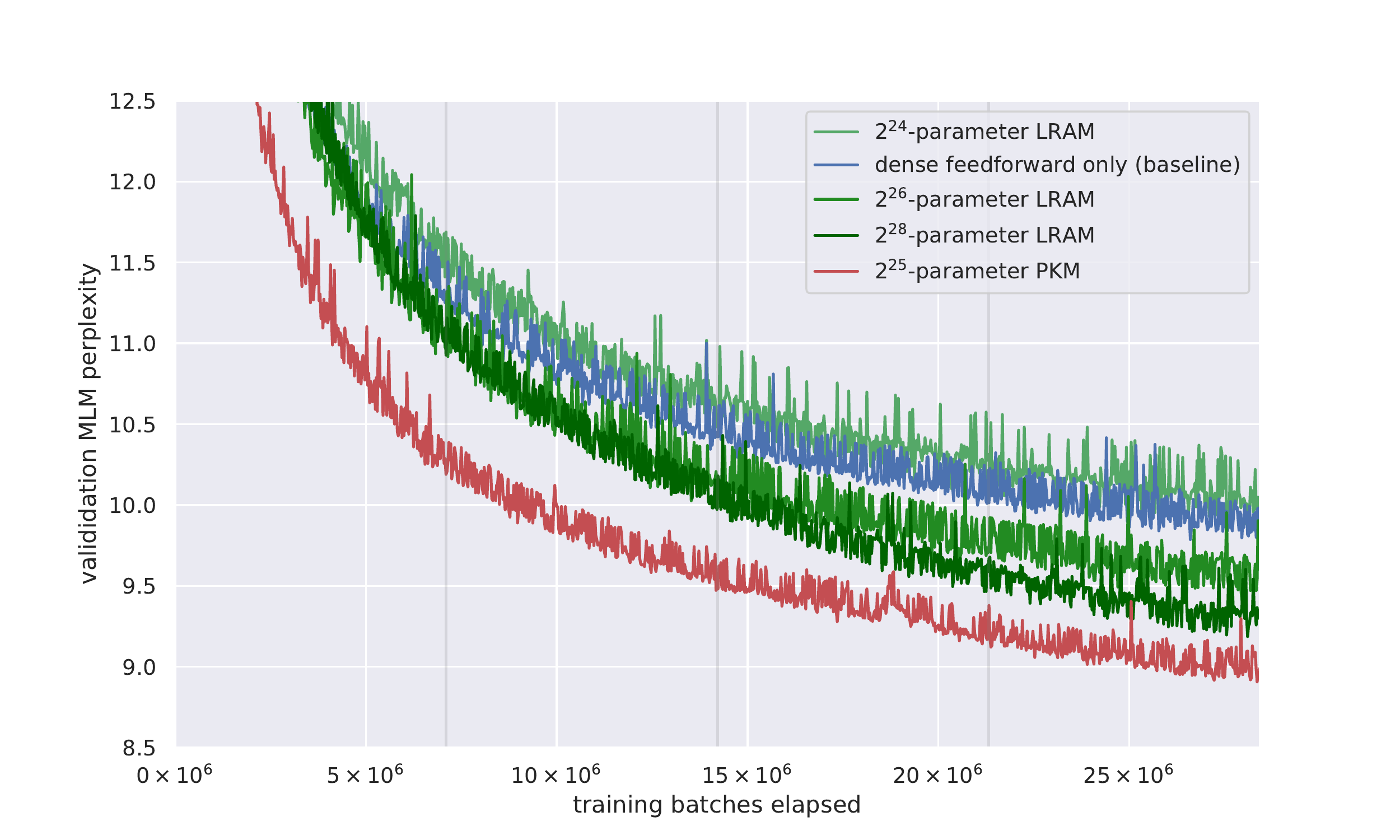}
  \caption{Validation perplexity over a period of 4 complete passes over the
training data, each shown as a vertical grey line. The training data consists of $227.4 \times 10^6$ paragraphs
and the batch size is 32 paragraphs.}
  \label{fig:validation}
\end{figure}

The best validation perplexity (from training for 4 complete passes over
the training data) and corresponding testing perplexity is tabulated in
Table~\ref{tab:ppl}.
The validation perplexity as a function of number of training iterations
is plotted in Figure~\ref{fig:validation}.

\begin{table}
  \caption{Validation and test perplexities for the language models.}
  \label{tab:ppl}
  \centering
  \begin{tabular}{c c c c}
    \toprule
    Model & Total parameters (Millions) & Validation perplexity & Test perplexity \\
    \midrule
    Baseline & 58.4 & 9.79 & 9.79 \\
    PKM & 90.3 & 8.92 & 8.92 \\
    LRAM-small & 74.4 & 9.89 & 9.87 \\
    LRAM-medium & 124.8 & 9.41 & 9.42 \\
    LRAM-large & 326.1 & 9.19 & 9.21 \\
    \bottomrule
  \end{tabular}
\end{table}

We see consistent improvement as we increase the memory capacity, with
negligible slowdown.

\subsection{Time Scaling}
\begin{figure}
  \centering
  \begin{subfigure}[b]{0.875\textwidth}
    \centering
    \includegraphics[width=\textwidth]{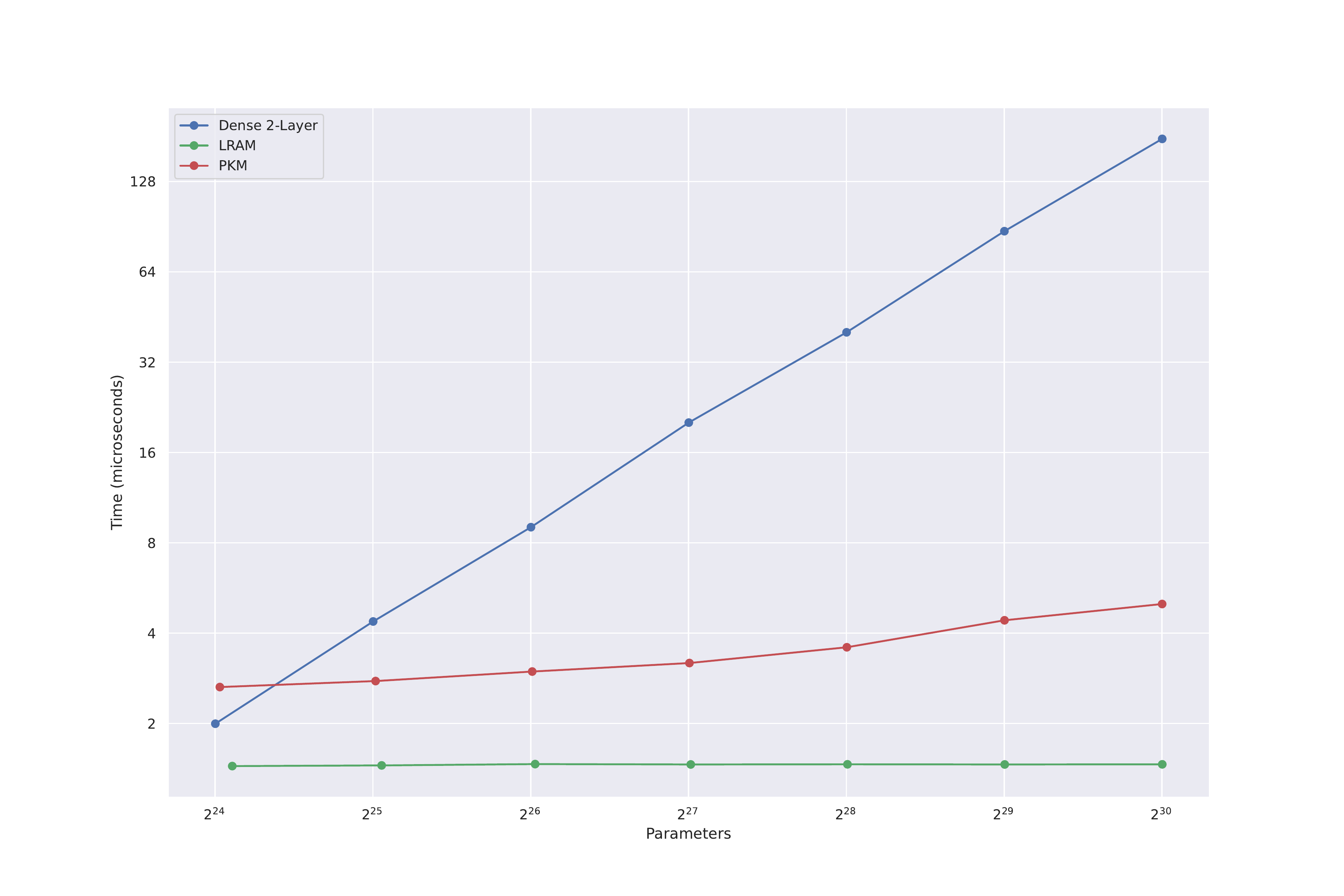}
    \caption{$w=512$}
  \end{subfigure}
  \begin{subfigure}[b]{0.875\textwidth}
    \centering
    \includegraphics[width=\textwidth]{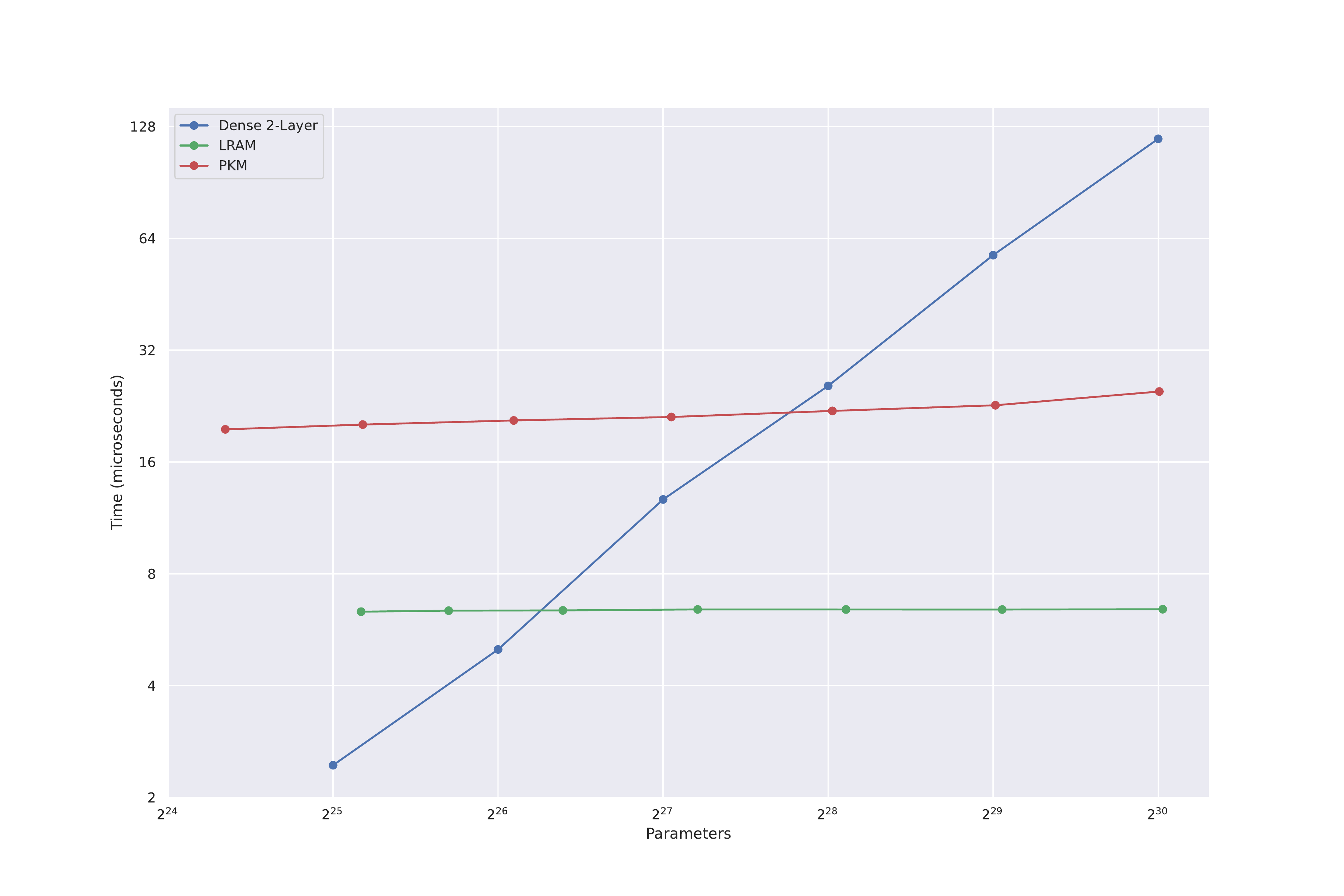}
    \caption{$w=2048$}
  \end{subfigure}
  \caption{Time for the forward pass through the layer, divided by the
  minibatch size, for each layer type as a function of total parameter
  count. Each time reported was the median of 15 successive runs.}
  \label{fig:timings}
\end{figure}

LRAM has excellent runtime performance and scalability in training and inference, which we tested to over a billion parameters in a single layer. Even at smallest tested size of approximately $2^{24}$ parameters, an LRAM layer runs 1.8x faster than PKM, with the advantage growing to 3.4x at the largest size tested. This advantage grows further at higher widths. Figure~\ref{fig:timings} shows the performance scaling of a simple dense 2-layer network, LRAM, and PKM at various widths.

\begin{table}
  \caption{Asymptotic scaling of layer types with width and parameter count. Here $r$ is the ratio of the hidden layer size to width, in this work always 4.}
  \centering
  \begin{tabular}{c c c}
    \toprule
    Method & Parameters & Approx Operation Count\\
    \midrule
    Dense 2-layer & $2rw^2$ & $2rw^2 + O(w)$ \\
    PKM & $mN + 2w\sqrt{N} + w^2$ & $2w\sqrt{N} + w^2 + O(w)$\\
    LRAM & $mN + \frac{5}{4}rw^2$ & $\frac{5}{4}rw^2 + O(w)$ \\
    \bottomrule
\end{tabular}
\end{table}

Although for the $w = 512$ transformers, the lattice memory layer remained
slower than the dense 2-layer network it replaces, at large enough widths
it becomes faster despite its increased parameter count. Table~\ref{tab:widths}
demonstrates that inference is slightly faster for LRAM when $w = 8192$, and
significantly faster for the $w = 12288$ used by GPT-3.\cite{gpt3}

\begin{table}
  \caption{Inference time per vector in microseconds as function of width. For LRAM, this remains essentially constant regardless of the memory size chosen.}
  \label{tab:widths}
  \centering
  \begin{tabular}{c c c}
  \toprule
  Width & Dense $w \rightarrow 4w \rightarrow w$ inference time & LRAM inference time \\
  \midrule
  2048 & 2.44 & 6.33 \\
  3072 & 5.57 & 10.64 \\
  4096 & 10.09 & 16.05 \\
  6144 & 27.58 & 31.28 \\
  8192 & 51.20 & 50.92 \\
 12288 & 124.3 & 106.2 \\
  \bottomrule
\end{tabular}
\end{table}

As shown in Figure~\ref{fig:timings}, as we scale parameter count, LRAM
cost stays essentially constant. LRAM has faster wall-clock times than
PKM across the board, at the cost of poorer perplexity at a given
parameter count.

\subsection{Memory Utilisation}
To check the memory usage patterns, we check both the fraction of memory locations accessed across the validation set, as well as the KL-divergence of the access from uniform as in \cite{pkm}. We find > 98\% utilisation for all sizes, with smaller memories having marginally higher. LRAM has comparable access uniformity to PKM with a similar number of memory locations, but as we use many more locations in the larger memories, we end up with less uniformity of access. The results are summarised in Table~\ref{tab:utilisation}.

\begin{table}
\caption{Proportion of memory accessed when performing MLM inference over
the validation dataset, along with the Kullback-Leibler divergence between
the weighted distribution of memory accesses and the uniform distribution.}
\label{tab:utilisation}
\centering
\begin{tabular}{c c c c c}
  \toprule
  Model & PKM & LRAM-small & LRAM-medium & LRAM-large \\
  \midrule
  Memory locations & $2^{16}$ & $2^{18}$ & $2^{20}$ & $2^{22}$ \\
  Parameters       & $2^{25}$ & $2^{24}$ & $2^{26}$ & $2^{28}$ \\
  \midrule
  Memory usage \% & 99.99 & 99.99 & 99.99 & 98.46 \\
  KL-divergence  & 1.57 & 1.57 & 1.64 & 2.52 \\
  \bottomrule
\end{tabular}
\end{table}

\section{Related Work}

PKM \cite{pkm} uses a Cartesian product structure to index in $O(\sqrt{N})$ time. This has the advantage that the lookup keys and queries can have arbitrary dimension, instead of the fixed 8-dimensional space LRAM looks up in. However, this comes at the cost of worse asymptotic scaling with memory size. In practice, we find that PKM outperforms LRAM for the same memory size, but has somewhat longer runtime in both training and inference. Based on the asymptotic scaling, we expect this runtime performance gap to increase with memory size, so depending on model hyperparameters one or the other may be preferable. 

SMoE \cite{smoe} maintains a collection of subnetworks, called experts, a fraction of which it selects on each forward pass. With appropriate hyperparameter choices, it can also achieve $O(\sqrt{N})$ scaling with the number of parameters. However, it requires additional care to avoid highly imbalanced expert selection. Due to implementation complexity, we do not compare against it directly.

\section{Conclusions and Future Work}
\label{section:conclusion}

We introduce a new type of differentiable memory layer using the $E_8$ lattice with $O(1)$ read and write regardless of size in a random access model, and show that in transformer-based masked language models it provides significant performance improvements, scaling with the memory capacity.

In \cite{pkm}, it was found that a deep transformer benefits from having two or more memory layers instead of one. The $O(1)$ complexity of LRAM means that instead of each of $\ell$ memory layers having separate sets of $N$ memory locations, it is no costlier to allow all $\ell$ layers to read from a shared set of $\ell N$ memory locations. This enables the trained model to adaptively control how much of the memory is used by each layer, as well as enabling parameter sharing.

LRAM could also be used as memory for an RNN, with both read and write capability every timestep. The differentiable neural computer (DNC) \cite{dnc} demonstrates the usefulness of this idea, but the reads and writes are dense (accessing every memory location) so computation time increases proportionally to the memory size. LRAM provides sparse operations, enabling a DNC to scale to vastly larger memory sizes without impacting the computation cost.

We chose the $E_8$ lattice due to known desirable properties, but these results open the possibility of trying higher-dimensional lattices at the cost of increasing the constant factor in the lookup complexity.

\section{Miscellaneous}
\label{section:misc}

The source code for the CUDA implementation of LRAM and PyTorch wrapper is
open-source under an MIT licence and available from \cite{latticelut}.

The experiments (language modelling and timing tests) were each performed on
a single NVIDIA\textregistered \; RTX 3090 GPU of a 4-GPU machine provided by
LeaderGPU \cite{leadergpu}. Each experiment took less than 3 GPU-months to
complete. The computing cost was funded by Hatsya Limited \cite{hatsya}.

As a general language model, any societal impacts would stem primarily from
the application of the language model and the dataset used to train it, rather
than from the architecture of the language model. Since uncurated datasets
automatically scraped from the Internet are typically replete with human
biases, and those biases may be reflected in any models trained on such data,
we have opted not to release the trained models.

% References begin here:

%%%%%%%%%%%%%%%%%%%%%%%%%%%%%%%%%%%%%%%%%%%%%%%%%%%%%%%%%%%%
\section*{Checklist}

\begin{enumerate}

\item For all authors...
\begin{enumerate}
  \item Do the main claims made in the abstract and introduction accurately reflect the paper's contributions and scope?
    \answerYes{}
  \item Did you describe the limitations of your work?
    \answerYes{See Section~\ref{section:conclusion}.}
  \item Did you discuss any potential negative societal impacts of your work?
    \answerYes{See Section~\ref{section:misc}.}
  \item Have you read the ethics review guidelines and ensured that your paper conforms to them?
    \answerYes{}
\end{enumerate}

\item If you are including theoretical results...
\begin{enumerate}
  \item Did you state the full set of assumptions of all theoretical results?
    \answerNA{}
  \item Did you include complete proofs of all theoretical results?
    \answerNA{}
\end{enumerate}

\item If you ran experiments...
\begin{enumerate}
  \item Did you include the code, data, and instructions needed to reproduce the main experimental results (either in the supplemental material or as a URL)?
    \answerYes{See section~\ref{section:misc}.}
  \item Did you specify all the training details (e.g., data splits, hyperparameters, how they were chosen)?
    \answerYes{See Subsections \ref{sub:hypers} and \ref{sub:dataset}.}
	\item Did you report error bars (e.g., with respect to the random seed after running experiments multiple times)?
    \answerNo{Retraining the neural networks with multiple random seeds would have been prohibitively costly.}
	\item Did you include the total amount of compute and the type of resources used (e.g., type of GPUs, internal cluster, or cloud provider)?
    \answerYes{See Section~\ref{section:misc}.}
\end{enumerate}

\item If you are using existing assets (e.g., code, data, models) or curating/releasing new assets...
\begin{enumerate}
  \item If your work uses existing assets, did you cite the creators?
    \answerYes{The XLM repository is cited, as are the natural language datasets in Subsection~\ref{sub:dataset}.}
  \item Did you mention the license of the assets?
    \answerYes{The code is released under an MIT licence as mentioned in Section~\ref{section:misc}.}
  \item Did you include any new assets either in the supplemental material or as a URL?
    \answerYes{The source code mentioned in Section~\ref{section:misc}.}
  \item Did you discuss whether and how consent was obtained from people whose data you're using/curating?
    \answerNA{}
  \item Did you discuss whether the data you are using/curating contains personally identifiable information or offensive content?
    \answerNA{}
\end{enumerate}

\item If you used crowdsourcing or conducted research with human subjects...
\begin{enumerate}
  \item Did you include the full text of instructions given to participants and screenshots, if applicable?
    \answerNA{}
  \item Did you describe any potential participant risks, with links to Institutional Review Board (IRB) approvals, if applicable?
    \answerNA{}
  \item Did you include the estimated hourly wage paid to participants and the total amount spent on participant compensation?
    \answerNA{}
\end{enumerate}

\end{enumerate}

%%%%%%%%%%%%%%%%%%%%%%%%%%%%%%%%%%%%%%%%%%%%%%%%%%%%%%%%%%%%

\end{document}